\documentclass{article}

\usepackage{microtype}
\usepackage{graphicx}
\usepackage{subfigure}
\usepackage{booktabs} 

\usepackage{hyperref}

\usepackage{epsfig} 

\usepackage{amsmath}

\usepackage{listings}
\usepackage{xcolor}
\usepackage{float}

\DeclareMathOperator*{\argmax}{arg\,max}

\usepackage{multirow}

\usepackage[accepted]{icml2021}

\begin{document}

\twocolumn[
\icmltitle{Dual Monte Carlo Tree Search}



\icmlsetsymbol{equal}{*}

\begin{icmlauthorlist}
\icmlauthor{Prashank Kadam}{neu}
\icmlauthor{Ruiyang Xu}{neu}
\icmlauthor{Karl Lieberherr}{neu}
\end{icmlauthorlist}

\icmlaffiliation{neu}{Khoury College of Computer Sciences, Northeastern University, Boston, Massachusetts}

\icmlcorrespondingauthor{Prashank Kadam}{kadam.pr@northeastern.edu}

\icmlkeywords{Machine Learning, ICML}

\vskip 0.3in
]



\printAffiliationsAndNotice{\icmlEqualContribution} 

\begin{abstract}
AlphaZero, using a combination of Deep Neural Networks and Monte Carlo Tree Search (MCTS), has successfully trained reinforcement learning agents in a tabula-rasa way. The neural MCTS algorithm has been successful in finding near-optimal strategies for games through self-play. However, the AlphaZero algorithm has a significant drawback; it takes a long time to converge and requires high computational power due to complex neural networks for solving games like Chess, Go, Shogi, etc. Owing to this, it is very difficult to pursue neural MCTS research without cutting-edge hardware, which is a roadblock for many aspiring neural MCTS researchers. In this paper, we propose a new neural MCTS algorithm, called Dual MCTS, which helps overcome these drawbacks. Dual MCTS uses two different search trees, a single deep neural network, and a new update technique for the search trees using a combination of the PUCB, a sliding-window, and the \(\epsilon\)-greedy algorithm. This technique is applicable to any MCTS based algorithm to reduce the number of updates to the tree. We show that Dual MCTS performs better than one of the most widely used neural MCTS algorithms, AlphaZero, for various symmetric and asymmetric games.
\end{abstract}

\section{Introduction}

\begin{equation*}
    \text{MB}(q,n)=\begin{cases}
      \text{True}, \quad \text{if } n=2\\
      \text{False}, \quad \text{if } n<2\vee (q=0\wedge n>2)\\
      \exists m\in[1...n-1]:\\\text{MB}(q-1,m)\wedge \text{MB}(q-1,n-m)
    \end{cases}       
\end{equation*}

Deepmind's AlphaGo \cite{alphago} was the first algorithm to be able to beat the human Go champion. AlphaGo uses a combination of Monte Carlo Tree Search (MCTS) and a Deep Neural Network (DNN). However, the algorithm had some game-specific configurations, which did not make it a generic algorithm that could be used for playing any game. Deepmind later came up the AlphaZero \cite{alpha0} which was a generic algorithm that could be used to play any game. Since then, a combination of MCTS and a DNN is the most widely used way to build game-playing programs. There is a need to balance accurate state estimation by the DNN and the number of simulations by the MCTS. Larger DNNs would be better for accurate evaluations but will take a toll on the cost of computation.

On the other hand, a smaller network could be used for faster evaluations and thus a larger number of MCTS simulations for the given amount of time. There is a need to find the optimum trade-off between the network's size and the number of MCTS simulations. One of the recently developed methods for this purpose is the Multiple Policy Monte Carlo Tree Search (MPV-MCTS) algorithm \cite{mpvmcts} which combines two Policy-Value Neural Networks (PV-NNs) of different sizes to retain the advantages of each network. The smaller network would perform a larger number of simulations on its tree to assign priorities to each state based on its evaluations. The more extensive network would then evaluate these states starting from the highest priority to achieve better accuracy. This algorithm shows a notable improvement in the convergence over games compared to AlphaZero, but the two-network configuration requires high run-time memory due to which the algorithm is difficult to run locally on average hardware.

Another problem that all the neural MCTS algorithms face is the large number of updates required during the backup phase of the tree. These updates increase exponentially as the tree depth increases. The computational time takes a big hit due to these updates, and there is a need to reduce the number of updates to the tree while keeping the values of each tree node highly accurate. We propose a technique that uses a combination of a sliding window and $\epsilon$-greedy search over a Polynomial Upper Confidence Tree (PUCT) \cite{puct}, for achieving this objective, thus reducing the time required for updates considerably. This technique can be applied to any MCTS-based algorithm as it optimizes over the core algorithm.

In this paper, we developed a novel algorithm that helps overcome the drawbacks of AlphaZero, MPV-MCTS, and MCTS and helps in accelerating the training speeds over various symmetric and asymmetric games. We show our improvements over AlphaZero and MPV-MCTS using Elo-rating, $\alpha$-rank, and training times, which are the most widely used metrics for evaluating games. We train our model for symmetric and asymmetric problems with different complexities and show that our model performance would improve considerably compared to other neural-MCTS algorithms as the game's state space increases.

\vspace{5mm}
\section{Background}



\subsection{AlphaZero}

 In a nutshell,AlphaZero uses a single neural network as the policy and value approximator. During each learning iteration, it carries out multiple rounds of self-plays. Each self-play runs several MCTS simulations to estimate an empirical policy at each state, then sample from that policy, take a move, and continue. After each round of self-play, the game's outcome is backed up to all states in the game trajectory. Those game trajectories generated during self-play are then be stored in a replay buffer, which is used to train the neural network.

In self-play, for a given state, the neural MCTS runs a given number of simulations on a game tree ,rooted at that state, to generate an empirical policy. Each simulation, guided by the policy and value networks, passes through 4 phases:
\begin{enumerate}
\item{SELECT: }
At the beginning of each iteration, the algorithm selects a path from the root (current game state) to a leaf (either a terminal state or an unvisited state) according to a predictor upper confidence boundary (PUCB) algorithm \cite{puct}. Specifically, suppose the root is $s_0$. The UCB determines a serial of states $\{s_0, s_1, ..., s_l\}$ by the following process:
\begin{equation}
    \begin{split}
    & a_{i} =\argmax_a\left[Q(s_{i},a)+c\pi_\theta(s_{i},a)\frac{\sqrt{\sum_{a'} N(s_{i},a')}}{N(s_{i},a)+1}\right]\\
    & s_{i+1} =\text{move}(s_{i},a_{i})
    \end{split}
    \label{eq:mcts-select}
\end{equation}
It has been proved in \cite{Grill2020MonteCarloTS} that selecting simulation actions using Eq.\ref{eq:mcts-select} is equivalent to optimize the empirical policy 
\begin{equation}
\begin{split}
& \hat{\pi}(s,a)=\frac{1+N(s,a)}{|A|+\sum_{a'} N(s,a')}
\end{split}
\end{equation}
where $|A|$ is the size of current action space, so that it approximate to the solution of the following regularized policy optimization problem:
\begin{equation}
\begin{split}
    \pi^*&=
    \argmax_\pi\left[Q^T(s,\cdot)\pi(s,\cdot)- \lambda KL[\pi_\theta(s,\cdot),\pi(s,\cdot)]\right]\\
    \lambda &= \frac{\sqrt{\sum_{a'} N(s_{i},a')}}{|A|+\sum_{a'} N(s,a')}
\end{split}
    \label{eq:mcts-opt}
\end{equation}
That also means that MCTS simulation is an regularized policy optimization \cite{Grill2020MonteCarloTS}, and as long as the value network is accurate, the MCTS simulation will optimize the output policy so that it maximize the action value output while minimize the change to the policy network.
\item{EXPAND: }
Once the selected phase ends at an unvisited state $s_l$, the state will be fully expanded and marked as visited. All its child nodes will be considered as leaf nodes during next iteration of selection.
\item{ROLL-OUT: }
The roll-out is carried out for every child of the expanded leaf node $s_l$. Starting from any child of $s_l$, the algorithm will use the value network to estimate the result of the game, the value is then backed up to each node in the next phase.
\item{BACKUP: }
This is the last phase of an iteration in which the algorithm updates the statistics for each node in the selected states $\{s_0, s_1, ..., s_l\}$ from the first phase. To illustrate this process, suppose the selected states and corresponding actions are
$$\{(s_0,a_0),(s_1,a_1),...(s_{l-1},a_{l-1}),(s_l,\_)\}$$ 
Let $V_\theta(s_i)$ be the estimated value for child $s_i$. We want to update the Q-value so that it equals to the averaged cumulative reward over each accessing of the underlying state, i.e., $Q(s,a)=\frac{\sum_{i=1}^{N(s,a)}\sum_tr_t^i}{N(s,a)}$. To rewrite this updating rule in an iterative form, for each $(s_t,a_t)$ pair, we have:
\begin{equation}
\begin{split}
    N(s_t,a_t)&\leftarrow N(s_t,a_t)+1\\
    Q(s_t,a_t)&\leftarrow Q(s_t,a_t)+\frac{V_\theta(s_r)-Q(s_t,a_t)}{N(s_t,a_t)}
\end{split}
    \label{eq:mcts-backup}
\end{equation}

Such a process will be carried out for all of the roll-out outcomes from the last phase.
\end{enumerate}
Once the given number of iterations has been reached, the algorithm returns the empirical policy $\hat{\pi}(s)$ for the current state $s$. After the MCTS simulation, the action is then sampled from the $\hat{\pi}(s)$, and the game moves to the next state. In this way, for each self-play iteration, MCTS samples each player's states and actions alternately until the game ends, which generates a trajectory for the current self-play. After a given number of self-plays, all trajectories will be stored into a replay buffer so that it can be used to train and update the neural networks.

\subsection{Multiple Policy Value Monte Carlo Tree Search}

AlphaZero uses a single DNN, called a Policy-Value Neural Network (PV-NN), consisting of two heads, one for the policy and one for value approximation.  In this subsection, we explain the Multiple Policy Value Monte Carlo Tree Search algorithm \cite{mpvmcts} in which the overall system consists of two PV-NNs $f_S$ and $f_L$ (smaller and the larger network, respectively). Let $b_{S}(b_L)$ be an assigned number of simulations or budget, the corresponding network used would be $f_{S}(f_L)$ this network would have its search tree $T_{S}(T_L)$. As a part of this problem, we would now like to find a stronger policy:
$$\pi(s,(f_{S}(b_S),(f_{L}(b_L)))$$
such that $b_{S} \geq b_L$. If the state is common to both the trees, $V(s)$ and prior probability $P(s, a)$ would be the same. For each simulation, we choose either $f_S$ or $f_L$ and depending on the value chosen. We further choose the leaf state of $T_S$ or$T_L$ to evaluate and update the tree. Any number of ways could be defined for the networks to take turns as long as $b_{S} \geq b_L$

This architecture's objective is for $T_S$ to provide the benefits of the look-ahead search where it balances between exploration and exploitation as it grows. Thus $f_S$, being a smaller network, should perform a higher number of simulations than $f_L$ with the same amount of resources. In each simulation for $f_S$, a leaf is selected, using the simulation count of $f_S$. In the case of $f_L$, for each simulation, we need to identify the critical states. Here we assume that the nodes in $T_S$ with higher visit counts are more important, and hence these states are assigned a higher value while evaluating $T_L$. To achieve this, we select the states with the highest visit counts from $T_S$ for each simulation. In case a particular leaf is not visited by $f_S$ (rare), we re-select an unevaluated leaf state using PUCB for $f_L$ instead. This process is repeated until convergence is achieved. Here the overall network is expected to have faster convergence due to the advantage of having a look ahead search using $T_S$.


\vspace{5mm}


\section{Dual Monte Carlo Tree Search (Dual MCTS)}

This section defines the neural MCTS algorithm that we have developed as a part of this paper, which we call the Dual MCTS algorithm. The idea behind this algorithm is similar to MPV-MCTS, where we have two different tree simulations, i.e., a smaller tree and a larger tree, and we want to provide a look-ahead to the larger tree by running a higher number of simulations on the smaller tree and prioritizing the states evaluated by the smaller tree. In our case, we have used a single neural network to simulate both the trees instead of using two separate networks like MPV-MCTS. We also introduce a novel method to reduce the number of value updates to the MCTS using a combination of the PUCT, a sliding window over the tree levels, and the $\epsilon$-greedy algorithm. This technique applies to any neural-MCTS based algorithm.

We have a single DNN that generates two different trees. This network would consist of two policy-value heads connected to one of the network's intermediate layers, and the other one is connected to the last layer of the network. What we are essentially doing here is subsetting a single network into two different sub-networks where one acts as a smaller network and the other acts as a larger network. We can consider the network from the input layer to the first policy-value head as the smaller network of $f_{sub}$ and the network from the input layer to the second policy-value head, i.e., the complete network as the larger network or $f_{full}$. Each of the networks $f_{sub}$ and $f_{full}$ generate their own trees, $T_{sub}$ and $T_{full}$. Since $f_{sub}$ is a smaller network, we can allocate a large simulation budget $b_{sub}$ to this network. Thus, $f_{sub}$ will generate large self-plays in a short period. Although this network's value estimations will not be very accurate, we only want to prioritize the most visited states. This priority list can then be used by $f_{full}$ to give a higher preference to states with higher priorities while simulating the search tree. The two networks can then be represented by $f_{sub}(f_{full})$ and the corresponding search trees can be represented by $T_{sub}(T_{full})$. The policy over which we would like to optimize can therefore be given by:
$$\pi(s,(f_{sub}(b_{sub}),(f_{full}(b_{full})))$$

such that $b_{sub} \geq b_{full}$. Where $b_{sub}$ and $b_{full}$ are budgets assigned to $f_{sub}$ and $f_{full}$ respectively. In case that we find a state which is common between both the tress, the policy, and the value for that state would be the same. Several methods can be used to combine multiple strategies during MCTS value evaluations. Some of these include Rapid Action Value Evaluation (RAVE) \cite{rave}, implicit minimax backups \cite{impbackup} and asynchronous policy value MCTS (APV-MCTS) \cite{asynmcts}. The first two combine MCTS evaluations with other heuristics. Meta MPV-MCTS is most related to Asynchronous Policy Value-MCTS \cite{asynmcts}. Hence, we will use the following method, which is similar to Asynchronous Policy Value-MCTS:

$$V(s) = \alpha V_{sub}(s) + (1-\alpha)V_{full}(s)$$

$$P(s,a) = \beta p_{sub}(a|s) + (1-\beta) p_{full}(a|s)$$

where $\alpha,\beta \in [0,1]$, are weight coefficients. Any of the two networks can go for the first simulation based on the approach of the implementation, suppose $f_{sub}$ is selected to go first, then we select a leaf from $T_{sub}$ to be evaluated and accordingly the values are updated for each of the states. The value of $b_{sub}$ and $b_{full}$ can be chosen conveniently. For this paper, we choose the value of $b_L$ as a random number between 1 and n and then choose $b_{sub}$ to be $\gamma b_{full}$, where $\gamma \in [1,1.5]$ and throughout multiple runs, we figure out which combination of $b_{sub}$ and $b_{full}$ fits the best for our combinatorial game.

We now introduce a novel way of updating the MCTS algorithm. Here, we use a combination of PUCT, a sliding window ($\tau$), and an $\epsilon$-greedy algorithm for action selection. During the MCTS search tree's backup phase, we update the values evaluated using the PUCT algorithm only up to the window $\tau$ previous nodes for the trajectory being followed in the current iteration. This method considerably reduces the number of updates required to be made to the MCTS, but this also impacts our PUCT algorithm's exploration term as $N(s, a)$ would not be updated for states outside our sliding window. For this purpose, we combine our sliding window approach with an $\epsilon$-greedy algorithm, which would help us facilitate exploration. The mathematical formalism of our action selection approach can be given as follows:





The PUCB action selection is defined as:
$$A_k = \argmax_a\left[Q(s_{i},a)+cP_\phi(a|s_{i})\frac{\sqrt{\sum_{a'} N(s_{i},a')}}{N(s_{i},a)+1} \right]$$

Here we add a sliding window to our PUCB algorithm such that the algorithm updates only the node upto window size ($\tau$) from the current node. The sliding window PUCB action selection will then occur in the following way:

\begin{equation*}
    \begin{cases}
      \forall 0 \leq k \leq \tau-1, \quad A_k = a_k\\
      \forall \tau \leq k \leq K-1, \quad A_k = \argmax_a [Q(s_{i},a, \tau)+ \\ \hspace{35mm} cP_\phi(a|s_{i})\frac{\sqrt{\sum_{a'} N(s_{i},a',\tau)}}{N(s_{i},a,\tau)+1}]
    \end{cases}       
\end{equation*}

where $a_k$ is the action evaluated for that node once the state is out of the window. Here we introduce the $\epsilon$-greedy algorithm in order to balance the exploitation and exploration outside the window $\tau$. As for within the window, the PUCB will take care of this. The $\epsilon$-greedy sliding window PUCB thus becomes:

\begin{equation*}
    \begin{cases}
      \forall 0 \leq k \leq \tau-1 \quad \& \quad U_k \geq \epsilon_{0}\nu^{k}, \quad A_k = a_k\\
      \forall 0 \leq k \leq \tau-1 \quad \& \quad U_k < \epsilon_{0}\nu^{k}, \quad A_k = Y_k\\
      
      \forall N \leq k \leq K-1 \quad , \quad A_k= \argmax_a [Q(s_{i},a, \tau)+ \\ \hspace{35mm} cP_\phi(a|s_{i})\frac{\sqrt{\sum_{a'} N(s_{i},a',\tau)}}{N(s_{i},a,\tau)+1}]
    \end{cases}       
\end{equation*}

Notice that we have introduced the term $\nu^{k}$, this is the decay term that is tuned for exploration depending on the state space of the game. The update for our $\epsilon$-greedy sliding-window PUCB then becomes:

\begin{equation*}
    \begin{cases}
      \forall 0 \leq k \leq \tau-1, \quad Q^{new}(s_{i}, a_{i}) = Q^{old}(s_{i}, a_{i}),\\
                                      \hspace{24mm} N^{new}(s_i,a_i)=N^{old}(s_i,a_i)\\
      \forall \tau \leq k \leq K-1,\\
      \hspace{12mm} Q^{new}(s_i,a_i)=\frac{Q^{old}(s_i,a_i)\times N^{old}(s_i,a_i)+V_{\phi}(s_r)}{N^{old}(s_i,a_i)+1}, \\
      \hspace{12mm} N^{new}(s_i,a_i)=N^{old}(s_i,a_i)+1
    \end{cases}       
\end{equation*}

\begin{figure}[ht]
\vskip 0.2in
\begin{center}
\centerline{\includegraphics[width=80mm]{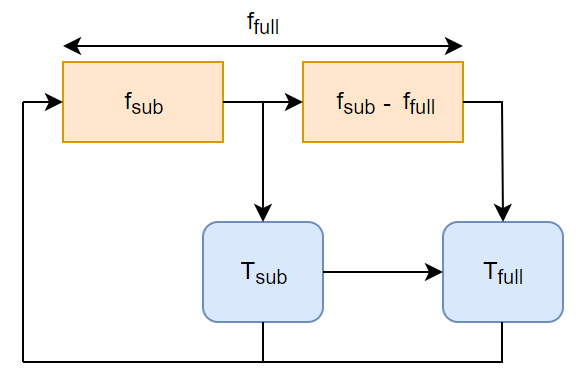}}
\caption{Dual MCTS block diagram}
\label{icml-historical}
\end{center}
\vskip -0.2in
\end{figure}


During each simulation for $T_{sub}$, the leaf state is selected based on our action selection algorithm, and for $T_{full}$ each state is selected based on the priorities assigned by $T_{sub}$ to each state, higher the state value, higher the priority. In case a particular state has not been visited by $T_{sub}$, the evaluation in $T_{full}$ is done using our action selection technique. The choice of $\tau$ for the sliding window is based on the state space of the game being evaluated. For smaller games like Nim, the value of $\tau$ would be small, but as the board size would increase, $\tau$ would also increase proportionally to the state space logarithmically.


\begin{algorithm}[tb]
  \caption{Dual MCTS}
  \label{alg:example}
\begin{algorithmic}
  \STATE $s\leftarrow Initial board state$
  \STATE $f_{full} \leftarrow Initialize the full network$
  \STATE $b_{full} \in [1,n],\quad b_{sub} \leftarrow \gamma b_{full}, \gamma \in [1,1.5]$
  \FOR{$i=1$ {\bfseries to} $b_{sub} + b_{full}$}
  \IF{$i \in [1, b_{sub}]$}
  \STATE $s_{leaf} = Select Unevaluated Leaf State (T_{sub})$;
  \STATE $(p,v) = f_{sub}(s_{leaf});$
  \STATE $Update(T_{sub}, s_{leaf}, (p,v));$
  \ELSE 
  \STATE $s_{leaf} = Select Unevaluated Leaf State by$
  \STATE \hspace{10mm} $Priority(T_{full})$;
  \IF{$N_{sub}(s_{leaf}) = 0$}
  \STATE $s_{leaf} = Select Unevaluated Leaf State(T_{full})$;
  \ENDIF 
  \STATE $(p,v) = f_{full}(s_{leaf});$
  \STATE $Update(T_{full}, s_{leaf}, (p,v));$
  \ENDIF
  \ENDFOR
  \caption{Dual MCTS}
  \label{mmm_alg}

\end{algorithmic}
\end{algorithm}


\vspace{5mm}

\section{Experiments}

This section describes our experiments to evaluate the Dual MCTS algorithm that we have defined above. The algorithm has been implemented in Python in Deepmind's OpenSpiel framework \cite{openspiel}. All the experiments have been performed on a core i7-7500U processor with 8 Gigabytes of RAM. For these experiments, we will use two problems: The first one is the single pile Nim problem, which is a symmetric problem where both the players will have the same action space. The second one is the Highest Safe Rung (HSR) \cite{hsr, xu_lieberherr_2020} which is an asymmetric problem which means that both the players have different action spaces. We claim that our algorithm performs better as the state space of the game increase. We also show our evaluations over the Connect-four game, which has a considerably larger state space than HSR or Nim. Since the Connect-four game took a long time to run on our current configuration, we upgraded the system to 16 Gigabytes of RAM. Note that this configuration is specific to the Connect-4 evaluations only.

The hyper-parameters used for the neural network are kept constant through all the different algorithms that we evaluate. In the case of the number of MCTS simulations, the number of simulations for smaller and the larger tree is kept the same in MPV-MCTS and Dual MCTS. The number of simulations applied to the AlphaZero search tree is the sum of the simulations of Dual MCTS's two search trees. For the DNN, we use ResNet with four layers and 64 neurons each. For Dual MCTS, $f_{sub}$ consists of the first two layers, and its output is connected to a policy-value head, and the complete network acts as the $f_{full}$. For MPV-MCTS, we have two separate networks, each having 4 and 6 layers for the smaller and the more extensive network, respectively. The number of simulations assigned to the smaller tree is 50, and that to the larger tree is 35.

\vspace{5mm}

\section{Evaluations}

We are going to evaluate each of the algorithms including AlphaZero, MPV-MCTS and Dual MCTS for HSR, Nim and Connect-4 using the following four metrics:

\begin{itemize}
    \item Elo Rating \cite{elo}
    \item $\alpha$-rank \cite{alpharank}
    \item Time per Iteration
    \item Time required for convergence
\end{itemize}

For the purpose of comparisons, we assume the convergence when the $\alpha$-rank score of the algorithm crosses 0.9. Note that this does not mean that the algorithm has for the optimal strategy to play the game but we use this metric to compare these algorithms which each other and it gives us a fair estimate of how these algorithms are performing.

\subsection{HSR Evaluations}

The Highest Safe Rung (HSR) \cite{algo} \cite{hsr,xu_lieberherr_2020} is an asymmetric problem in which both the agents' action spaces are different. HSR(k,q,n) basically defines a stress testing problem, where one, given $k$ jars and $q$ test chances, throwing jars from a specific rung of a given ladder with height $n$ to locate the highest safe rung. If $n$ is appropriately large, then one can locate the highest safe rung with at most $k$ jars and $q$ test times; otherwise, if $n$ is too big, then there is no way to locate the highest safe rung. This problem can be described as following:
$$HSR(k,q,n)=\begin{cases} \mbox{True},\ \mbox{if } n=1 \\ \mbox{False},\ \mbox{if } n>1\wedge (k=0\vee q=0) 
\\\exists m\in [1..n]:
HSR(k-1,q-1,m)\\\wedge HSR(k,q-1,n-m)
\end{cases}$$


It is well known that the above formulae can be translated into a logic game using Hintikka rules. \cite{sep-logic-games}.

\begin{figure}[ht]
\vskip 0.2in
\begin{center}
\centerline{\includegraphics[width=80mm]{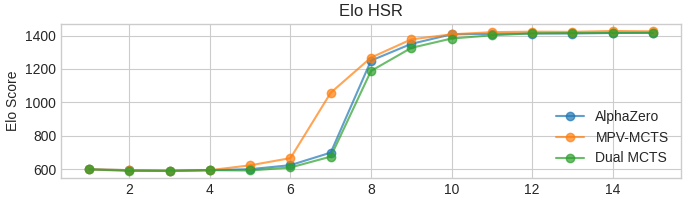}}
\centerline{\includegraphics[width=80mm]{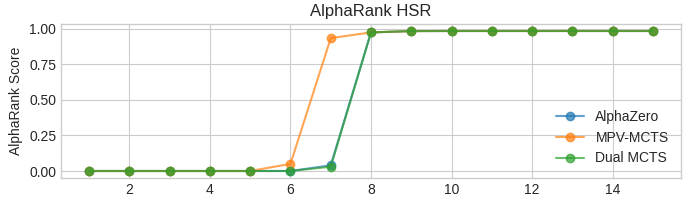}}
\centerline{\includegraphics[width=80mm]{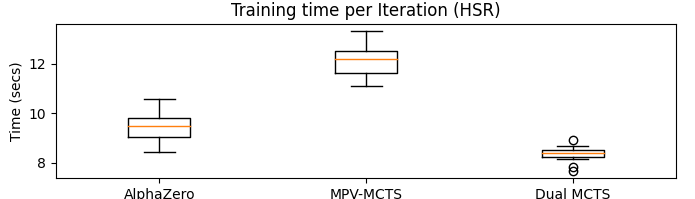}}
\centerline{\includegraphics[width=80mm]{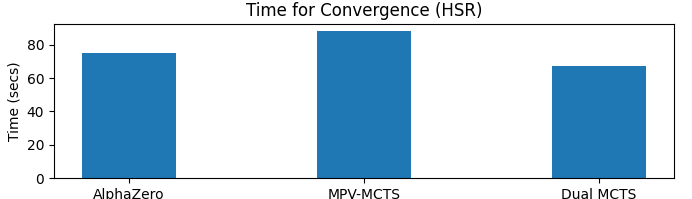}}
\caption{HSR Evaluations}
\label{hsr}
\end{center}
\vskip -0.2in
\end{figure}

    

Here we use HSR(4,4,16), which means that the board size is (1x16) with four tests available and four questions that could be asked. Figure \ref{hsr} shows how each of the algorithms performs for the HSR problem. As we see from Figure \ref{hsr}, MPV-MCTS converges a step faster than AlphaZero and Dual MCTS for both Elo-rating as well as the $\alpha$-rank. However, the time required per training step for AlphaZero and Dual MCTS is much lesser than MPV-MCTS, due to which the total time taken for convergence of MPV-MCTS is much more than AlphaZero or Dual MCTS, we see that Dual MCTS outperforms AlphaZero by 11.58\%. Dual MCTS outperforms MPV-MCTS by 29.42\%.

\subsection{Nim Evaluations}

\begin{figure}[ht]
\vskip 0.2in
\begin{center}
\centerline{\includegraphics[width=80mm]{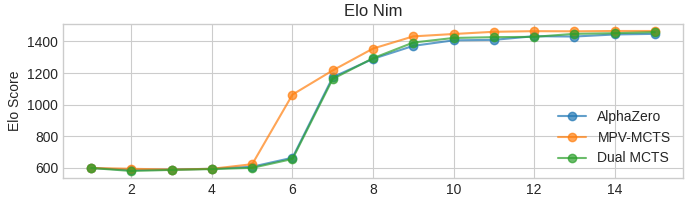}}
\centerline{\includegraphics[width=80mm]{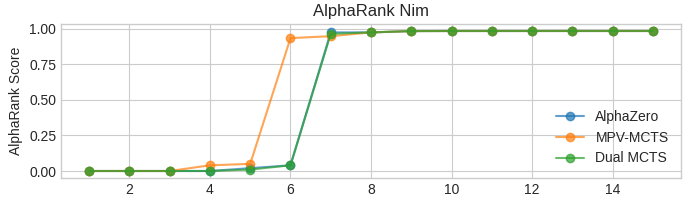}}
\centerline{\includegraphics[width=80mm]{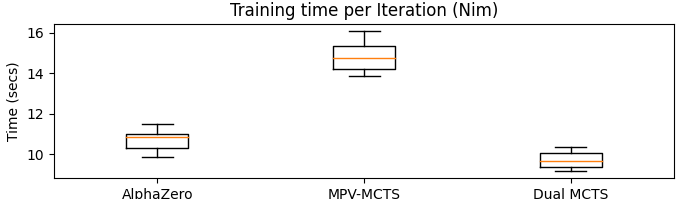}}
\centerline{\includegraphics[width=80mm]{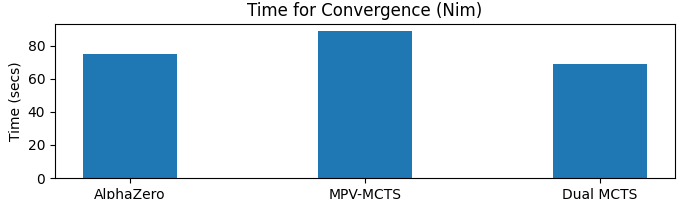}}
\caption{Nim Evaluations}
\label{nim}
\end{center}
\vskip -0.2in
\end{figure}

Nim \cite{Nim} is a symmetric problem, where each player is allowed to choose $x$ stones from a pile of $n$ stones turn by turn where $x \leq n$. The player to pick the last stone wins. Figure \ref{nim} the results of our evaluations on the Nim(3,20) game

Here we see quite similar results to the HSR problem, where although the MPV-MCTS converges a step faster than AlphaZero or Dual MCTS, it takes a much longer time per each training iteration due to its dual network configuration. Dual MCTS, in this case, surpasses AlphaZero by 8.84\% and MPV-MCTS by 27.78\% faster training.

\subsection{Connect 4 Evaluations}

\begin{figure}[ht]
\vskip 0.2in
\begin{center}
\centerline{\includegraphics[width=80mm]{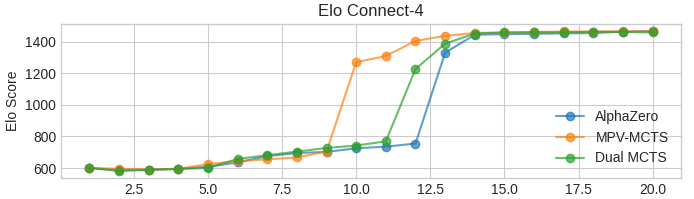}}
\centerline{\includegraphics[width=80mm]{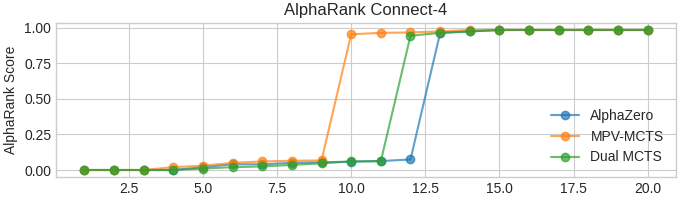}}
\centerline{\includegraphics[width=80mm]{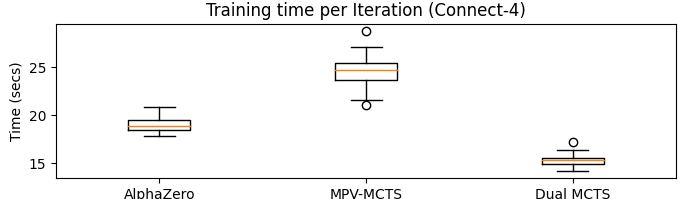}}
\centerline{\includegraphics[width=80mm]{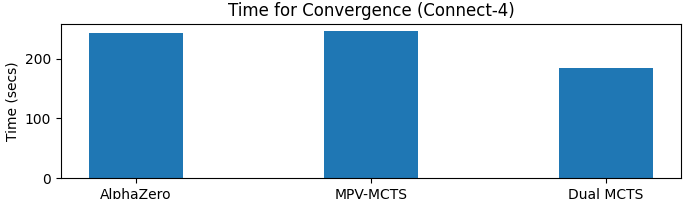}}
\caption{Connect-4 Evaluations}
\label{conn_4}
\end{center}
\vskip -0.2in
\end{figure}

To show that our algorithm outperforms the more conventional algorithms like AlphaZero and MPV-MCTS by even larger margins, we evaluate the game of Connect-4 \cite{connect}. Connect-4 has a much larger state space than HSR(4,4,16) or Nim (3,20). Figure \ref{conn_4} shows the results of our evaluations for all the models that we are testing.

As we can see from Figure \ref{conn_4}, Dual MCTS performs much better than AlphaZero or MPV-MCTS. As we have seen from our previous evaluations, there is a similar pattern where MPV-MCTS takes a lower number of timesteps to converge than AlphaZero or Dual MCTS, but it is taking a much longer time per iteration. As a result, Dual MCTS outperforms AlphaZero by 32.30\% in training time and MPV-MCTS by 33.92\% in training time. This result provides a piece of conclusive evidence that our algorithm will work much better for problems with larger state spaces.

Note that throughout all the experiments which we have performed, Dual MCTS constantly outperforms AlphaZero and MPV-MCTS in terms of the convergence time. All these experiments are run on very modest hardware and we show here that Dual MCTS could be a very useful algorithm for use-cases where high-end hardware is unavailable. Another thing to notice here is how both MPV-MCTS and Dual MCTS perform across various state spaces. MPV-MCTS converges much faster for Connect-4 than for HSR/Nim as compared to the other algorithms but takes more time per training step as it has to update two different neural networks. In the very same, Dual MCTS also shows faster convergence for Connect-4 game, this shows us the advantage of multiple tree searches. Along with this a single network configuration and an efficient update strategy help Dual MCTS converge much faster than its competitors.

\begin{table*}[!htbp]
\caption{Table showing the details of the average time required per training step, no. of steps required for convergence and the total time for convergence for each of the algorithms}
\begin{center}
\begin{tabular}{ |p{2cm}||p{1cm}|p{1cm}|p{1cm}||p{1cm}|p{1cm}|p{1cm}||p{1cm}|p{1cm}|p{1cm}|}
\hline
\multicolumn{10}{|c|}{\textbf{Evaluations}} \\
\hline
\multirow{2}{*}{\textbf{Algorithm}} & \multicolumn{3}{c||}{\textbf{HSR}} & \multicolumn{3}{c||}{\textbf{Nim}} & \multicolumn{3}{c|}{\textbf{Connect-4}}\\\cline{2-10}
& \textbf{Time (Step)}  & \textbf{Steps (Conv.)} & \textbf{Time (Conv.)} & \textbf{Time (Step)}  & \textbf{Steps (Conv.)} & \textbf{Time (Conv.)} & \textbf{Time (Step)}  & \textbf{Steps (Conv.)} & \textbf{Time (Conv.)}\\
\hline
\textbf{AlphaZero} & 9.34  & 8 & 74.72 & 10.71 & 7 & 74.97 & 18.71 & 13 & 243.23\\
\hline
\textbf{MPV-MCTS} & 12.38 & 7 & 86.66 & 14.67 & 6 & 88.02 & 24.62 & 10 & 246.20\\
\hline
\textbf{Dual MCTS} & 8.37 & 8 & 66.96 & 9.84 & 7 & 68.88 & 15.32 & 12 & 183.84\\
\hline
\end{tabular}
\end{center}
\end{table*}

\vspace{5mm}

\section{Conclusions}

In this paper we have proposed a new neural MCTS algorithm, Dual MCTS which uses a combination of two MCTSs along with an novel update technique for the Monte Carlo Search Tree to reduce the number of updates to the tree. This whole contraption accelerates the agent training and we train our agents for different symmetric and asymmetric problems using Dual MCTS and show that Dual MCTS performs better than more conventionally used algorithms like AlphaZero \cite{alpha0} and MPV-MCTS \cite{mpvmcts}. We also show how Dual MCTS shows even better improvements in convergence times as the state space of the problems that we solve, increases. We show our evaluations using different metrics.

Dual MCTS can also be scaled up for larger problems by creating smaller intermediate trees ($>2$), which could provide a look-ahead to the following tree in a chained fashion and help achieve even faster convergence in huge state spaces. Establishing a constraint which can accurately form such hierarchies and form a balance between the state space of the problem and the number of trees required for the fastest solution is a part of our future work.


\bibliography{example_paper}
\bibliographystyle{icml2021}





\end{document}